\documentclass[conference]{IEEEtran}
\IEEEoverridecommandlockouts

\usepackage{cite}
\usepackage{amsmath,amssymb,amsfonts}
\usepackage{algorithmic}
\usepackage{graphicx}
\usepackage{textcomp}
\usepackage{xcolor}

\usepackage{times}
\usepackage{soul}
\usepackage{url}
\usepackage[hidelinks]{hyperref}
\usepackage[utf8]{inputenc}
\usepackage[small]{caption}
\usepackage{graphicx}
\usepackage{booktabs}
\usepackage{algorithm}
\usepackage[switch]{lineno}

\usepackage{multirow}

\def\BibTeX{{\rm B\kern-.05em{\sc i\kern-.025em b}\kern-.08em
    T\kern-.1667em\lower.7ex\hbox{E}\kern-.125emX}}
\begin{document}

\title{scAGC: Learning Adaptive Cell Graphs with Contrastive Guidance for Single-Cell Clustering}


\author{\IEEEauthorblockN{Huifa Li\textsuperscript{1,2}, Jie Fu\textsuperscript{3}, Xinlin Zhuang\textsuperscript{2}, Haolin Yang\textsuperscript{2}, Xinpeng Ling\textsuperscript{4}, Tong Cheng\textsuperscript{1}, \\Haochen Xue\textsuperscript{2}, Imran Razzak\textsuperscript{2*}, Zhili Chen\textsuperscript{1*}\thanks{* Corresponding author: Imran Razzak, Zhili Chen.}}
\IEEEauthorblockA{\textit{\textsuperscript{1}Shanghai Key Laboratory of Trustworthy Computing, East China Normal University, Shanghai, China}}
\IEEEauthorblockA{\textit{\textsuperscript{2}Department of Computational Biology, Mohamed bin Zayed University of AI, Abu Dhabi, UAE}}
\IEEEauthorblockA{\textit{\textsuperscript{3}Department of Computer Science, Stevens Institute of Technology, Hoboken, USA}}
\IEEEauthorblockA{\textit{\textsuperscript{4}College of Electronics and Information Engineering, Tongji University, Shanghai, China}}
\IEEEauthorblockA{Huifa.Li@mbzuai.ac.ae}
}

\maketitle

\begin{abstract}

Accurate cell type annotation is a crucial step in analyzing single-cell RNA sequencing (scRNA-seq) data, which provides valuable insights into cellular heterogeneity. However, due to the high dimensionality and prevalence of zero elements in scRNA-seq data, traditional clustering methods face significant statistical and computational challenges. While some advanced methods use graph neural networks to model cell-cell relationships, they often depend on static graph structures that are sensitive to noise and fail to capture the long-tailed distribution inherent in single-cell populations.
To address these limitations, we propose scAGC, a single-cell clustering method that learns adaptive cell graphs with contrastive guidance. Our approach optimizes feature representations and cell graphs simultaneously in an end-to-end manner. Specifically, we introduce a topology-adaptive graph autoencoder that leverages a differentiable Gumbel-Softmax sampling strategy to dynamically refine the graph structure during training. This adaptive mechanism mitigates the problem of a long-tailed degree distribution by promoting a more balanced neighborhood structure. To model the discrete, over-dispersed, and zero-inflated nature of scRNA-seq data, we integrate a Zero-Inflated Negative Binomial (ZINB) loss for robust feature reconstruction. Furthermore, a contrastive learning objective is incorporated to regularize the graph learning process and prevent abrupt changes in the graph topology, ensuring stability and enhancing convergence. Comprehensive experiments on 9 real scRNA-seq datasets demonstrate that scAGC consistently outperforms other state-of-the-art methods, yielding the best NMI and ARI scores on 9 and 7 datasets, respectively.
Our code is available at \href{https://anonymous.4open.science/r/2025_BIBM_scAGC-5EF3}{Anonymous Github}.

\end{abstract}

\section{Introduction}

Single-cell RNA sequencing (scRNA-Seq) technology has become a driving force in the analysis of the cellular heterogeneity of tissues, which provides researchers valuable insights into critical biological topics, such as the discovery of rare/novel cell types \cite{jindal2018discovery}, the study of complex diseases like the detection of cancer stem cells \cite{bhaduri2020outer,paik2020single}, the elucidation of cell heterogeneity \cite{papalexi2018single}, the exploration of complex systems \cite{stubbington2017single} and the inference of cell trajectory \cite{saelens2019comparison}. Accurate cell type identification is an important step in the single-cell analysis mentioned above. Clustering has been proven to be an effective method for defining cell types in an unsupervised manner \cite{kiselev2019challenges}. However, although many classic clustering algorithms such as K-means \cite{macqueen1967some} and spectral clustering are robust and universal for traditional clustering tasks, clustering analysis of scRNA-seq data remains a statistical and computational challenge due to its unique high-dimensional characteristics and have a large number of zero elements \cite{Li2020DeepLE,Eraslan2018SinglecellRD}.
\begin{figure}[!t]
    \centering
    \includegraphics[width=1.0\linewidth]{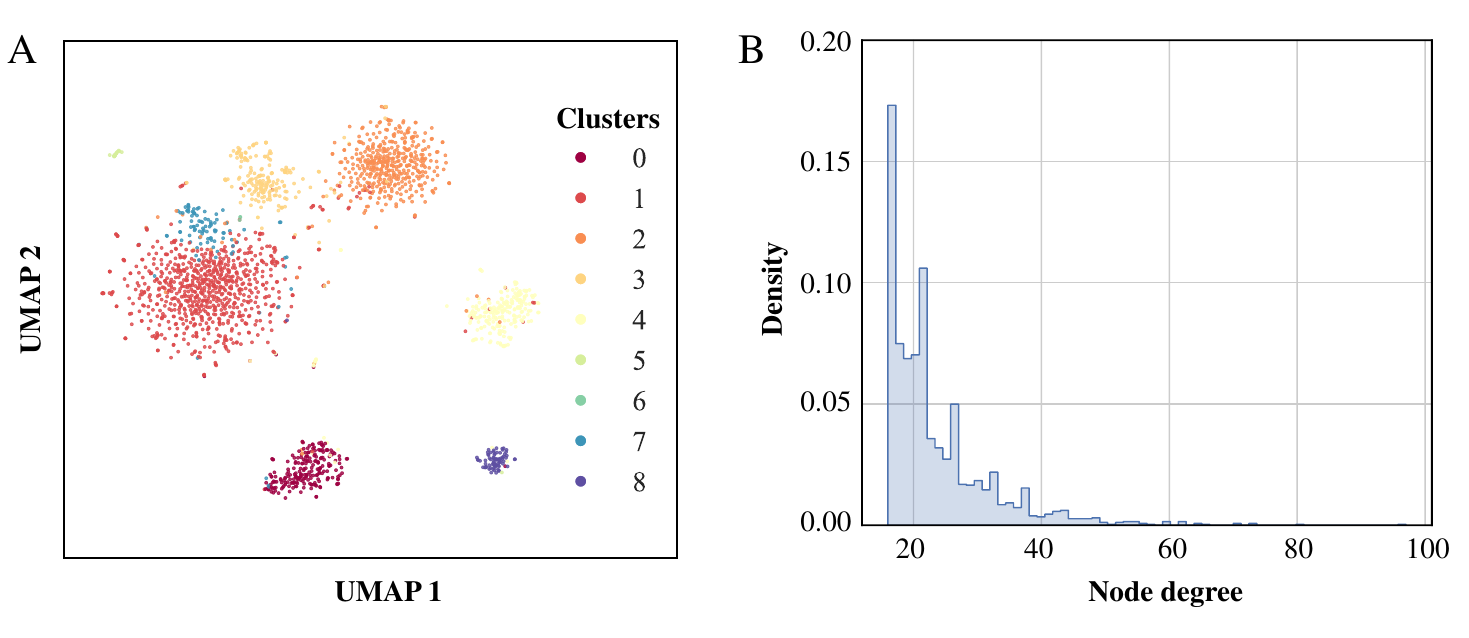}
    \caption{Visualization of the intrinsic long-tail structure in the Muraro single-cell dataset. (A) This visualizes the dataset in a UMAP-reduced space, which clearly departs from a uniform distribution, highlights the scale-free nature of cell-cell similarity networks. (B) The long-tailed degree distribution of the K-Nearest Neighbor (KNN) graph shows that most cells connect to only a few others, while a minority of cells act as highly connected hubs.}
    \label{fig:intro}
\end{figure}

In recent years, deep embedding clustering approaches have emerged as a promising paradigm for modeling the high-dimensional and sparse scRNA-seq data such as scDC \cite{tian2019clustering}, scziDesk \cite{chen2020deep}, scDCC \cite{tian2021model} and scGMAI \cite{yu2021scgmai}. 
These methods often integrate zero-inflated negative binomial (ZINB) distributions to effectively capture the intrinsic properties of gene expression data, including discreteness, zero inflation, and over-dispersion. Clustering is typically performed in a learned latent space using divergence-based objectives, such as Kullback–Leibler (KL) divergence. Despite their success, most of these frameworks focus solely on modeling individual cell profiles while neglecting the relational structure among cells, which limits their ability to learn robust and biologically meaningful representations.

\begin{figure*}[!t]
    \centering
    \includegraphics[width=1.05\linewidth]{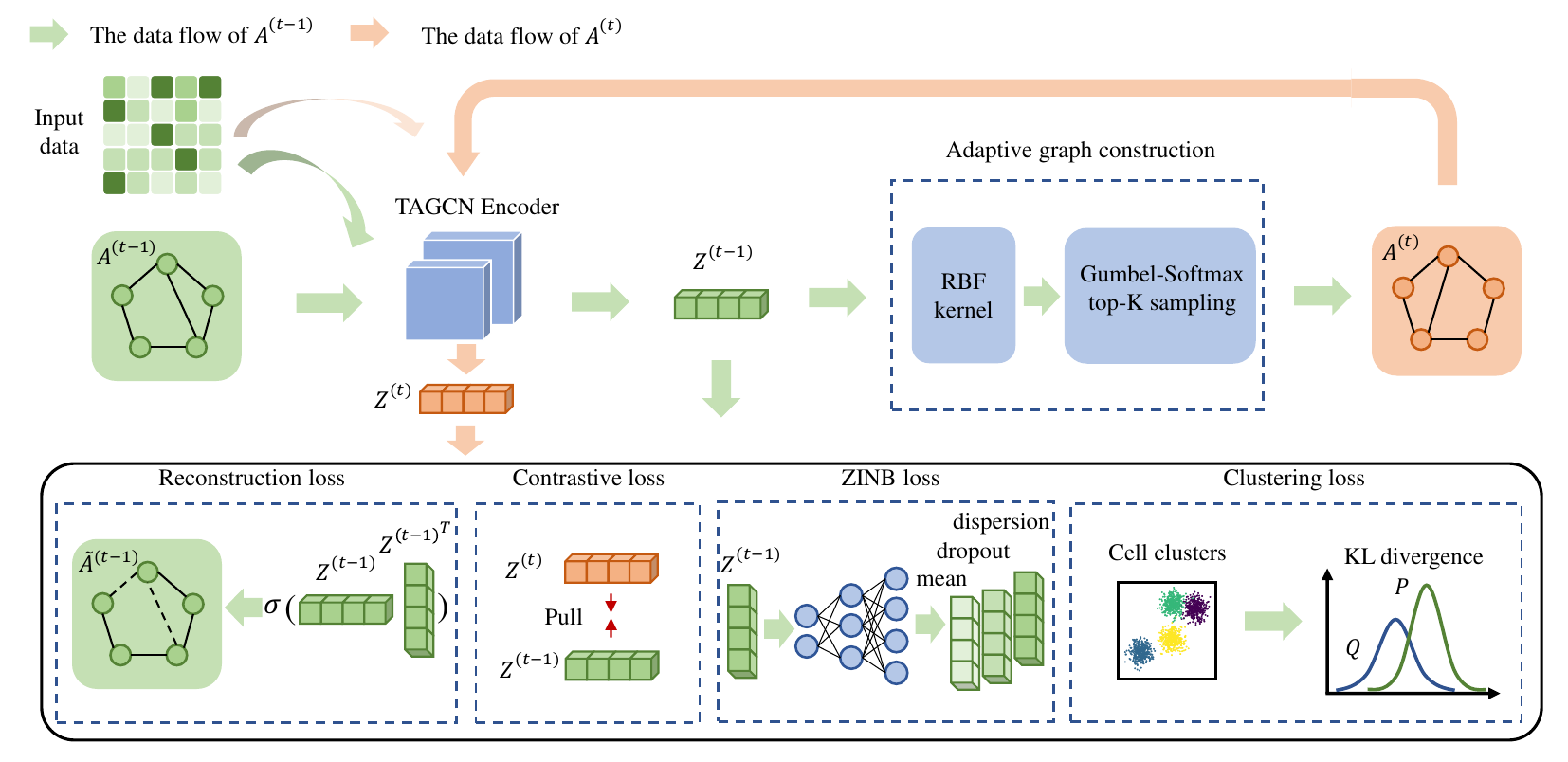}
    \caption{Overview of scAGC. Overview of the proposed framework. At epoch $t$, we construct adaptive cell graphs end-to-end using an adial basis function (RBF) kernel and Gumbel-Softmax sampling. A contrastive loss regularizes the evolution between graphs $A_{t-1}$ and $A_t$, while ZINB loss models data sparsity and guides topology reconstruction. Clustering is optimized via KL divergence.}
    \label{fig:graph_autoencoder}
\end{figure*}
Recently, emerging graph neural networks (GNNs) based models such as scGAE \cite{luo2021topology}, scGNN \cite{wang2021scgnn}, scTAG \cite{yu2022zinb} and scE2EGAE \cite{wang2025sce2egae} have been proposed to address the limitations of deep clustering methods by explicitly incorporating cell–cell relational information. 
These methods typically leverage graph autoencoders to model the topological structure of cellular graphs, thereby learning latent representations that are more suitable for clustering.
While these GNN-based methods have demonstrated notable improvements in single-cell clustering performance, two critical challenges remain. 1) They often depend on static graph structures derived from filtered gene expression features. Such fixed graphs are inherently sensitive to noise in the input data and may fail to faithfully capture the underlying cellular architecture. 2) The intrinsic long-tailed distribution of single-cell populations gives rise to graph topologies, particularly those constructed via $K$-nearest neighbor (KNN) algorithms, which exhibit highly skewed degree distributions as illustrated in Fig.~\ref{fig:intro}. In these graphs, a small number of supernodes dominate connectivity, potentially creating information bottlenecks and exacerbating the over-smoothing problem during GNN training.
Collectively, these challenges reflect the structural sparsity and imbalance pervasive in biological cell graphs, which can severely hinder uniform message-passing mechanisms. Overcoming these limitations remains a central challenge for advancing GNN-based clustering in the single-cell domain.

\noindent  \textbf{Our work.} To address the challenges of graph construction and representation learning in single-cell clustering, we propose scAGC, an end-to-end framework that adaptively learns cell graphs guided by clustering objectives, as illustrated in Fig. \ref{fig:graph_autoencoder}. Specifically, we first construct an initial cell graph using the KNN algorithm based on highly variable genes. To enable differentiable and task-adaptive graph learning during training, we replace the hard top-$K$ neighbor selection with a Gumbel-Softmax sampling strategy \cite{Jang2016CategoricalRW}, which assigns probabilities to all potential neighbors and softly samples $K$ connections. This serves as the foundation of a topology-adaptive graph autoencoder, which jointly learns dynamic graph structures and informative cell representations. Notably, this process also implicitly transforms the degree distribution from long-tailed to bell-shaped, improving graph regularity.
Moreover, to model the inherent sparsity and zero inflation in scRNA-seq data, we incorporate a ZINB loss for robust reconstruction. In addition, we introduce a contrastive learning objective to regularize graph evolution and promote stable, consistent topological relationships throughout training. For the downstream clustering task, we employ KL divergence to optimize clustering assignments, enabling the model to produce well-separated and biologically coherent clusters.

\noindent \textbf{Contribution.} Compared to previous methods, the contributions of our method are summarized as follows:
\begin{itemize}
    \item We propose scAGC, an end-to-end framework that jointly optimizes feature representations and cell graphs using a differentiable Gumbel-Softmax sampling strategy, effectively mitigating long-tailed degree distributions.
    \item We introduce a contrastive objective regularizes graph evolution during training, preventing abrupt topological shifts and improving convergence.
    \item We incorporate a Zero-Inflated Negative Binomial loss to model the sparsity, discreteness, and overdispersion of scRNA-seq data.
    \item The comprehensive empirical experiment results demonstrate that scAGC outperforms the other baselines on most of the 9 real scRNA-seq datasets.
\end{itemize}

\section{Related Work}

We divide the state-of-the-art methods for clustering scRNA-seq data into two main categories: deep clustering methods, and deep graph clustering methods. 

The deep embedding methods perform embedding learning and clustering simultaneously. In \cite{tian2019clustering}, scDC integrates the ZINB model to learn cell representation, while the clustering task on latent space is performed by clustering with Kullback–Leibler (KL) divergence. In \cite{chen2020deep}, scziDesk combines the deep learning technique with the use of a denoising autoencoder to characterize scRNA-seq data while proposing a soft self-training K-means algorithm to cluster the cell population in the learned latent space. In \cite{tian2021model}, scDCC integrates prior domain knowledge into the clustering process by utilizing soft pairwise constraints, which are derived from known relationships between cells. In \cite{yu2021scgmai}, scGMAI utilizes autoencoder networks to reconstruct gene expression values from scRNA-Seq data and FastICA is used to reduce the dimensions of reconstructed data.

Deep graph clustering methods supply the deep clustering framework with a message-passing mechanism to preserve the structural information during the encoding process. 
In \cite{luo2021topology}, scGAE builds a cell graph and uses a multitask‑oriented graph autoencoder to preserve topological structure information and feature information in scRNA‑seq data simultaneously. In \cite{wang2021scgnn}, scGNN formulates and aggregates cell-cell relationships with graph neural networks and models heterogeneous gene expression patterns using a left-truncated mixture Gaussian model. In \cite{yu2022zinb}, scTAG integrates the ZINB model into a topology adaptive graph convolutional autoencoder to learn the low dimensional latent representation and adopts KL divergence for the clustering tasks. In \cite{hu2023scdfc}, scDFC proposes a novel single-cell deep fusion clustering model to make full use of attribute information of each cell or the structure information between different cells. scE2EGAE \cite{wang2025sce2egae} constructs cell graphs in an end-to-end manner to denoise the input data, but it lacks explicit optimization for the clustering objective, resulting in suboptimal clustering performance.

\section{Preliminary and Motivation}

\subsection{Problem Definition}
In this study, we use the preprocessed scRNA-seq gene expression data to construct a cell graph. 
Let $\mathcal{G} = (\mathcal{V},\mathcal{E}, \mathbf{X})$ denote a cell graph, where $\mathcal{V}=\{v_1,v_2,\dots,v_N\}$, $\mathcal{E} \subseteq \mathcal{V} \times \mathcal{V}$ represent the node set and the edge set respectively.
We denote the feature matrix as $\mathbf{X} \in \mathbb{R}^{N \times M}$, where $\mathbf{X}_{ij}$ denotes the expression count of $j$-th highly-variable gene in $i$-th cell.
We denote the adjacency matrix as $\mathbf{A} \in \{0,1\}^{N \times N}$, where $\mathbf{A}_{i,j}=1$ if cell $i$ is cell $j$’s neighbor within top-$K$ shortest Euclidean distance.
Our goal is to divide a group of cells into several clusters through optimized biological signals obtained from the scRNA-seq data.

\subsection{Cell Graph based KNN Algorithm}
We leverage embeddings learned from a graph autoencoder to preserve both relational and neighborhood information among units. Following prior work, we construct unit graphs using the K-nearest neighbors (KNN) algorithm, where each node corresponds to a unit. Specifically, for any two nodes $a$ and $b$, a directed edge from $a$ to $b$ is added if $b$ is among the $k$-nearest neighbors of $a$, with $K$ set to 15. Pairwise correlations are measured using Euclidean distance to identify the top-$K$ nearest neighbors for each node. The resulting KNN graph is initially directed; we symmetrize it by adding reciprocal edges, yielding an undirected graph with binary edge weights uniformly set to 1.

\section{Methods}

\subsection{Adaptive Graph Learning via Gumbel-Softmax }
To enable end-to-end learning of a task-specific cell graph, we propose a differentiable neighbor selection mechanism based on Gumbel-Softmax sampling. Given the node embeddings $Z \in \mathbb{R}^{N \times D}$ obtained from a GNN encoder, we first compute a pairwise similarity matrix $S \in \mathbb{R}^{N \times N}$. Specifically, we adopt an RBF kernel to measure similarity:
\begin{equation}
    S_{ij} = \exp \left( -\frac{\| z_i - z_j \|^2}{2\sigma^2} \right)
\end{equation}
where $z_i$ and $z_j$ are the embedding vectors of nodes $i$ and $j$, and $\sigma$ is a bandwidth hyperparameter. However, directly selecting the top-$K$ neighbors for each row of $S$ is non-differentiable and thus incompatible with gradient-based optimization. To address this, we employ a Gumbel-TopK approximation. Specifically, we inject Gumbel noise $G_{ij} \sim \text{Gumbel}(0,1)$ into the similarity matrix to obtain a perturbed $S'_{ij}$:
\begin{equation}
    S'_{ij} = S_{ij} + G_{ij}
\end{equation}

To obtain a "soft" adjacency matrix that is differentiable, we apply a softmax operation with temperature $\tau$ over each row of $S'$:

\begin{equation}
    y_{ij} = \frac{\exp(S'_{ij}/\tau)}{\sum_{l=1}^{N} \exp(S'_{il}/\tau)}
\end{equation}


To enforce sparsity and retain discrete top-$K$ neighbor selection during forward computation, we apply a Straight-Through (ST) estimator. Specifically, we retain the top-$K$ hard  neighbors in the forward pass and use the soft distribution $y$ for gradient computation:
\begin{equation}
    A_{ij} = \text{stop\_gradient}(z_{ij}^{\text{hard}} - y_{ij}) + y_{ij}
\end{equation}
where $z^{\text{hard}}_{ij} = 1$ if node $j$ is among the top-$K$ neighbors of node $i$, and 0 otherwise. This results in an adaptive adjacency matrix $A \in \mathbb{R}^{N\times N}$ that is sparse and dynamically updated during training, allowing the model to learn graph structures.

To satisfy the input requirement of GNNs, which typically operate on undirected graphs, we convert the directed adjacency matrix A into a symmetric form. Specifically, if either $A_{ij}=1$ or $A_{ji}=1$, we set both $A_{ij}=A_{ji}=1$, ensuring that the resulting graph is undirected:
\begin{equation}
    A = \min(A + A^{T}, 1)
\end{equation}
This symmetrization step aligns the graph structure with the assumptions of standard GNN architectures and enables effective message passing between nodes.

This differentiable neighbor selection mechanism enables dynamic refinement of the graph structure during training, guided by downstream objectives such as representation learning and clustering. Moreover, it implicitly penalizes over-connected nodes in long-tailed degree distributions, promoting a more balanced and uniform neighborhood structure across the graph structure.

\subsection{Topology-Aware Graph Autoencoder with ZINB Loss}
To capture the precise intrinsic structure of cell features and cell-cell relationships in the cell graph, we develop a topology-aware graph autoencoder (TAGAE) with ZINB loss. Although most GNNs are allowed in our framework, we choose TAGCN \cite{kipf2016semi} as the base encoder, due to its comparable performance in node classification tasks. 

The standard GCN aggregates only the features of first-order neighbors at each layer, while TAGCN captures information from multi hop neighbors within a single layer by using a set of parallel graph convolution kernels ranging in size from 0 to $K$. This allows the model to adaptively learn the contribution of neighbors at different distances to the representation of the current node. The propagation rule of a TAGCN layer can be defined as:
\begin{equation}
    H^{(l+1)} = \sigma\left(\sum_{k=0}^{K} (\hat{D}^{-1/2} \hat{A} \hat{D}^{-1/2})^k H^{(l)} W_k^{(l)}\right)
\end{equation}
where $H^{(l)}$ is the node representation matrix of the $l$-th layer, and $H^{(0)}=X$. $\hat{A}=A+I$ is an adjacency matrix with self loops added, $\hat{D}$ is its corresponding degree matrix. $(\hat{D}^{-1/2} \hat{A} \hat{D}^{-1/2})^k$ represents an aggregator of $k$-order neighbors. $W_k^{(l)}$ is the learnable weight matrix corresponding to the $k$-th order neighbor aggregation in the $l$-th layer. $K$ is a hyperparameter that defines the maximum size of the convolution kernel. $\sigma$ is a non-linear activation function, such as ReLU.
By stacking $L$ layers of TAGCN, we ultimately obtain the potential representation of the node $Z=H^{(L)}$.

For cell-cell relationship learning, considering the intrinsic correlations of cells are preserved in the latent embedding space. We define the decoder of the graph autoencoder by an inner product, as formulated below:
\begin{equation}
    Z = f_E(\mathcal{G}),
    \widetilde{A} = \sigma(Z^TZ) \label{eq:rec_a}
\end{equation}
where $\widetilde{A}$ are the reconstructed adjacency matrix of the cell graph $\mathcal{G}$; $\sigma(\cdot)$ is the sigmoid activation function. Therefore, the reconstruction loss should be minimized in the learning process, we have the loss functions as below:
\begin{equation}
    \mathcal{L}_{g} = || A - \widetilde{A} ||_2^2
\end{equation}

For cell feature learning, considering the scRNA-seq gene expression data matrix exhibits characteristics of discrete, over-dispersed, and zero-inflated. 
We use a single decoder based on the zero-inflated negative binomial (ZINB) model to simultaneously model these characteristics, achieving the capture of the global probabilistic structure of the scRNA-seq data. The ZINB-based autoencoder is trained to attempt to reconstruct its input is defined as follows: 
\begin{align} 
    \mathrm{NB}(X|\mu,\theta) &= \frac{\Gamma(X+\theta)}{X!\Gamma(\theta)}(\frac\theta{\theta+\mu})^\theta(\frac\mu{\theta+\mu})^X \\
    \mathrm{ZINB}(X|\pi,\mu,\theta) &= \pi\delta_0(X)+(1-\pi)\mathrm{NB}(X)
\end{align}
where $\mu$ and $\theta$ represent the mean and dispersion in the negative binomial distribution, respectively; $\pi$ is the weight of the zero components. 
The ZINB-based decoder is a three layers fully connected neural network to estimate the parameters of ZINB.
Then, the reconstruction loss function of the original data $\mathbf{X}$ is defined as the negative log-likelihood of the ZINB distribution:
\begin{align}
    \mathcal{L}_{\mathrm{ZINB}} = -\log (\mathrm{ZINB}(X|\pi, \theta, \mu))
\end{align}

\begin{algorithm}[tb]
\caption{scAGC algorithm}
    \label{alg:algorithm}
    \textbf{Input}: Cell graph $\mathcal{G}=(\mathcal{V},\mathcal{E},X)$, the topology graph autoencoder TAGAE($f:\{f_E,f_D\}$), pre-training epochs $T_1$, formal-training epochs $T_2$ \\
    \ \textbf{Output}: TAGAE parameters $f$
    \begin{algorithmic}[1] 
    \STATE \textbf{\# Phase 1: pre-training}.
    \STATE Initialize $\mathcal{G}^{(0)} = \mathcal{G}$
    \FOR{$t_1$ $\leftarrow$ 1,2,$\dots$, $T_1$}
        \STATE Obtain node embeddings $Z^{(t_1-1)}$ of cell graph $\mathcal{G}^{(t_1-1)}$ using the encoder $f_E$ \label{algo:ob_emb}
        \STATE Obtain new cell graph $\mathcal{G}^{(t_1)}$ using the Gumbel-Softmax sampling strategy 
        \STATE Obtain node embeddings $Z^{(t_1)}$ of new cell graph $\mathcal{G}_t$ using the encoder $f_E$
        \STATE Obtain reconstructed $\widetilde{A}^{(t_1-1)}, \pi, \theta$ and $\mu$ of $\mathcal{G}$ using the decoder $f_D$ \label{algo:decode}
        \STATE Compute the embedding objective $\mathcal{L}_e$ with Eq.(\ref{eq:l_e})
        \STATE Update parameters by applying stochastic gradient ascent to minimize $\mathcal{L}_e$
    \ENDFOR
    \STATE \textbf{\# Phase 2: formal-training}.
    \STATE Let $\mathcal{G}^{(0)} = \mathcal{G}^{(T_1)}$
    \FOR{$t_2$ $\leftarrow$ 1,2,$\dots$,$T_2$}
        \STATE Same operation as \ref{algo:ob_emb}-\ref{algo:decode} in pre-training stage
        \STATE Obtain label distribution $Q$ with Eq.(\ref{eq:q})
        \STATE Compute the clustering objective $\mathcal{L}_c$ with Eq.(\ref{eq:l_c})
        \STATE Update parameters by applying stochastic gradient ascent to minimize $\mathcal{L}_c$
    \ENDFOR
    \STATE \textbf{return} TAGAE parameters $f$
\end{algorithmic}
\end{algorithm}

\subsection{Contrastive Guidance and Clustering Objective}
To enhance the stability of the learning graph structure and optimize for clustering tasks, we propose a contrastive contrastive loss and a clustering alignment loss.
During training, the adaptive adjacency matrix $A^{(t)}$ is updated iteratively. To prevent drastic changes between consecutive iterations and encourage structural consistency, we adopt a temporal contrastive loss that penalizes divergence between successive adjacency matrices.
Specifically, given two consecutive adjacency matrices $A^{(t-1)}$ and $A^{(t)}$, we use a shared graph encoder $f_E$ to compute the corresponding latent node embeddings:
\begin{equation}
    z_1 = f_E(X, A^{(t-1)}), \quad z_2 = f_E(X, A^{(t)})
\end{equation}

We treat $(z_1,z_2)$ as a positive pair, since they represent the same data $X$ under slightly different graph structures. Other samples in the batch serve as negative pairs. The contrastive loss encourages the two embeddings to remain close in latent space:
\begin{equation}
    \mathcal{L}_{contrast} = - \log \frac{\exp(\text{sim}(z_1, z_2) / \tau_c)}{\sum_{z' \in \mathcal{B} \setminus \{z_1\}} \exp(\text{sim}(z_1, z') / \tau_c)}
\end{equation}
where $sim(z_i,z_j) = \frac{z_j^T z_j}{||z_i|| ||z_j||}$ is cosine similarity; $\tau_c$ is the temperature coefficient that adjusts the sensitivity of positive and negative sample similarity; $\mathcal{B}$ is a comparison pool that includes all sample pairs. The final loss is averaged over all positive pairs $\mathcal{P}$:
\begin{equation}
    \mathcal{L}_{cg} = \frac{1}{|\mathcal{P}|} \sum_{(i,j) \in \mathcal{P}} \mathcal{L}_{contrast}(z_i, z_j)
\end{equation}
This term guides the model to evolve the graph topology smoothly over time, reducing instability and enhancing convergence in structure-sensitive tasks.

In parallel, we introduce a clustering objective that encourages node representations to form well-separated clusters in latent space. Specifically, we follow a soft assignment approach using a Student’s t-distribution and define a target distribution $Q$ based on the embedding similarity:
\begin{align}
    q_{ic}=\frac{(1+\left\|z_i-\mu_c\right\|^2)^{-1}}{\sum_r(1+\left\|z_i-\mu_r\right\|^2)^{-1}} \label{eq:q}
\end{align}
where $q_{ic}$ is the soft label of the embedding node $z_i$. This label measures the similarity between $z_i$ and the cluster central embedding $\mu_c$ by a Student’s t-distribution.
In addition, $p_{ic}$ is the auxiliary target distribution, which puts more emphasis on the similar data points assigned with high confidence on the basis of $q_{ic}$, as below:
\begin{align}
    p_{ic}=\frac{q_{ic}^2/\sum_iq_{ic}}{\sum_r(q_{ir}^2/\sum_iq_{ir})}
\end{align}
The clustering loss is then defined via KL divergence between the target distribution $P$ and the soft assignment $Q$:
\begin{equation}
    L_{KL}=KL(P||Q) 
    =\sum_i\sum_c p_{ic}log\frac{p_{ic}}{q_{ic}}
\end{equation}
This objective explicitly sharpens cluster assignments while preserving embedding smoothness.

\subsection{Training Strategy}
The overall training process of scAGC contains two stages, namely, the embedding learning stage and the cluster assignment stage. In the embedding learning stage, the loss function for training is defined as:
\begin{align}
    \mathcal{L}_e = \lambda_1 \mathcal{L}_{g} + \lambda_2 \mathcal{L}_{\mathrm{ZINB}} + \lambda_3 \mathcal{L}_{cg} \label{eq:l_e}
\end{align}
In the cluster assignment stage, the loss function for training is defined as:
\begin{align}
    \mathcal{L}_c = \lambda_1 \mathcal{L}_{g} + \lambda_2 \mathcal{L}_{\mathrm{ZINB}} + \lambda_3 \mathcal{L}_{cg} + \lambda_4 \mathcal{L}_{KL} \label{eq:l_c}
\end{align}
where $\lambda_1,\lambda_2,\lambda_3$ and $\lambda_4$ are the weight coefficients for each loss. The detailed description of our framework is provided in Algorithm \ref{alg:algorithm}.

\section{Experiments}
\begin{table}[htbp]
    \caption{Summary of the nine real scRNA-seq datasets.}
    \centering
    \resizebox{1.0\columnwidth}{!}{
    \begin{tabular}{ccccccc}
        \toprule
        \makebox[0.05\columnwidth][c]{Dataset} & \makebox[0.020\columnwidth][c]{Cells} & \makebox[0.045\columnwidth][c]{Genes} & \makebox[0.015\columnwidth][c]{Class} & \makebox[0.1\columnwidth][c]{Platform} &\makebox[0.1\columnwidth][c]{Reference}\\
        \midrule
        QS\_Diap & 870 & 23341 & 5 & Smart-seq2 & \cite{schaum2018single} \\
        QS\_LM & 1090 & 23341 & 6 & Smart-seq2 & \cite{schaum2018single} \\
        QS\_Lung & 1676 & 23341 & 11 & Smart-seq2 & \cite{schaum2018single} \\
        Muraro & 2122 & 19046 & 9 & CEL-seq2 & \cite{muraro2016single} \\
        Adam  & 3660 & 23797 & 8 & Drop-seq & \cite{adam2017psychrophilic} \\
        QX\_LM & 3909 & 23341 & 6 & 10x & \cite{schaum2018single} \\
        QS\_Heart & 4365 & 23341 & 8 & Smart-seq2 & \cite{schaum2018single} \\
        Young & 5685 & 33658 & 11 & 10x & \cite{young2018single} \\
        Plasschaert & 6977 & 28205 & 8 & inDrop & \cite{plasschaert2018single} \\
        Wang\_Lung & 9519 & 14561 & 2 & 10x & \cite{wang2018pulmonary} \\
        \bottomrule
    \end{tabular}}
    \label{tab:datasets}
\end{table}
\subsection{Data Sources and Preprocessing}

In our experiments, we use 9 real scRNA-seq datasets collected from \cite{yu2022zinb} to demonstrate the effectiveness of scAGC. These datasets are from different species and originate from five different representative sequencing platforms. The cell numbers range from 870 to 9519, and the cell type numbers vary from 2 to 11. The detailed information is described in Table \ref{tab:datasets}. 
In the preprocessing stage, we first remove genes with zero expression across all cells. Next, we normalize the expression matrix $X_{ij}$ by dividing each entry by the total expression of cell $i$, followed by a natural logarithm transformation to obtain continuous log-scale values. We then select the top $M$ highly variable genes based on normalized variance rankings by the Scanpy package. Finally, we construct a cell graph using the KNN algorithm, where each node represents a cell. For each cell, edges are established to its top $K$ most similar neighbors based on expression similarity.

\subsection{Baseline}
We select 10 other state-of-the-art single-cell clustering methods for comparison, as listed below:
\begin{itemize}
    \item Deep embedding methods: scDC \cite{tian2019clustering}, scziDesk \cite{chen2020deep}, scDCC \cite{tian2021model} and scGMAI \cite{yu2021scgmai}.
    \item Deep Graph embedding methods: scGAE \cite{luo2021topology}, scGNN \cite{wang2021scgnn}, scTAG \cite{yu2022zinb} and scE2EGAE \cite{wang2025sce2egae}.
    \item Traditional methods: K-means and Spectral clustering.
\end{itemize}

\begin{table*}[htbp]
    \caption{Performance comparison between various baselines on 9 real datasets.} 
    \centering
    \resizebox{1.0\textwidth}{!}{
    \begin{tabular}{c|c|c|cccc|cccc|cc}
        \toprule
            \makebox[0.04\textwidth][c]{\multirow{2}{*}{Metric}} & \multirow{2}{*}{Dataset} & \makebox[0.05\textwidth][c]{Ours} & \multicolumn{4}{c|}{Deep Graph Clustering} & \multicolumn{4}{c|}{Deep Clustering} & \multicolumn{2}{c}{Base Clustering} \cr
            \cmidrule{3-13}
            & & \makebox[0.04\textwidth][c]{scAGC} & \makebox[0.06\textwidth][c]{scE2EGAE} & \makebox[0.04\textwidth][c]{scTAG} & \makebox[0.04\textwidth][c]{scGAE} & \makebox[0.04\textwidth][c]{scGNN} & \makebox[0.04\textwidth][c]{scziDesk} & \makebox[0.03\textwidth][c]{scDC} & \makebox[0.04\textwidth][c]{scDCC} & \makebox[0.04\textwidth][c]{scGMAI} & \makebox[0.04\textwidth][c]{Kmeans} & \makebox[0.04\textwidth][c]{Spectral} \\
        \midrule
        \multirow{9}{*}{NMI}
         & \makebox[0.06\textwidth][c]{QS\_Diap} & \textbf{0.9417} & 0.8437 & 0.8929 & 0.7351 & 0.7608 & 0.9210 & 0.7807 & 0.8223 & 0.6836 & 0.8846 & 0.8881 \\
         & QS\_LM & \textbf{0.9485} & 0.7375 & 0.9327 & 0.7398 & 0.7726 & 0.9468 & 0.7048 & 0.4624 & 0.7198 & 0.8911 & 0.9389 \\
         & QS\_Lung & \textbf{0.8115} & 0.7069 & 0.7845 & 0.6766 & 0.6642 & 0.7543 & 0.6840 & 0.4982 & 0.7312 & 0.7785 & 0.7976 \\
         & Muraro & \textbf{0.8695} & 0.8034 & 0.8000 & 0.7619 & 0.6294 & 0.7349 & 0.7549 & 0.8347 & 0.7168 & 0.8194 & 0.8266 \\
         & Adam & \textbf{0.8852} & 0.6321 & 0.8651 & 0.6330 & 0.5587 & 0.8476 & 0.7680 & 0.7494 & 0.6163 & 0.0722 & 0.0998 \\
         & QX\_LM & \textbf{0.9509} & 0.9371 & 0.9273 & 0.5670 & 0.7166 & 0.9407 & 0.7359 & 0.8798 & 0.7639 & 0.4481 & 0.8443 \\
         & QS\_Heart & \textbf{0.9281} & 0.7157 & 0.8689 & 0.6039 & 0.6540 & 0.8723 & 0.6531 & 0.4242 & 0.6941 & 0.8299 & 0.8454 \\
         & Plasschaert & \textbf{0.8621} & 0.6113 & 0.6046 & 0.5563 & 0.5856 & 0.4070 & 0.4070 & 0.5786 & 0.5711 & 0.6915 & 0.5216 \\
         & Wang\_Lung & \textbf{0.8580} & 0.6604 & 0.7150 & 0.3150 & 0.3975 & 0.7965 & 0.1511 & 0.5862 & 0.3432 & 0.7167 & 0.0367 \\
         \cmidrule{1-13}
         \multirow{9}{*}{ARI}
         & QS\_Diap & \textbf{0.9737} & 0.8683 & 0.9137 & 0.5638 & 0.5646 & 0.9517 & 0.6479 & 0.8895 & 0.4111 & 0.9110 & 0.9170 \\
         & QS\_LM & 0.9688 & 0.5334 & 0.9560 & 0.5419 & 0.6399 & \textbf{0.9743} & 0.5384 & 0.3449 & 0.4899 & 0.8922 & 0.9615 \\
         & QS\_Lung & 0.7376 & 0.4594 & 0.6252 & 0.2792 & 0.3631 & 0.7401 & 0.4504 & 0.2908 & 0.4622 & 0.7329 & \textbf{0.7559} \\
         & Muraro & \textbf{0.9158} & 0.7042 & 0.8050 & 0.6413 & 0.5080 & 0.6784 & 0.6609 & 0.7100 & 0.5132 & 0.8452 & 0.6436 \\
         & Adam & \textbf{0.8973} & 0.5223 & 0.8713 & 0.2490 & 0.4125 & 0.8431 & 0.6241 & 0.6576 & 0.4316 & 0.0218 & 0.0368 \\
         & QX\_LM & \textbf{0.9719} & 0.9677 & 0.9500 & 0.1300 & 0.6206 & 0.9127 & 0.4381 & 0.8023 & 0.5061 & 0.3775 & 0.8981 \\
         & QS\_Heart & \textbf{0.9621} & 0.5523 & 0.9299 & 0.2497 & 0.5222 & 0.9324 & 0.4673 & 0.2584 & 0.4368 & 0.8376 & 0.8757 \\
         & Plasschaert & \textbf{0.9192} & 0.4002 & 0.6203 & 0.0370 & 0.5856 & 0.4867 & 0.4070 & 0.4668 & 0.5711 & 0.7352 & 0.2916 \\
         & Wang\_Lung & \textbf{0.9274} & 0.7384 & 0.8014 & 0.1035 & 0.1771 & 0.8975 & 0.2520 & 0.5998 & 0.1325 & 0.7995 & 0.0345 \\
       \bottomrule
    \end{tabular}}
    
    \label{tab:cluster_result}
\end{table*}
\subsection{Implementation Details}
The proposed scAGC is constructed with Tensorflow 2.9.0, Spektral 1.3.1, and Python 3.8.10. 
In the proposed scAGC method, we constructed the cell graph with the KNN algorithm with the nearest neighbor parameter $K = 15$ and the number of highly variable genes $M=1500$. 
In graph autoencoder, each layer was configured with 512, 256, and 128 nodes; and the layer of the fully connected decoder was configured with a symmetric encoder form. 
Our algorithm consisted of pre-training and formal-training, which are set to 1000 and 300 epochs, respectively. 
All parameters were optimized by the Adam optimizer. For the optimizer, we set the learning rate of 1e-2 for pre-training and 5e-4 for formal training
For weight coefficients of contrastive learning module, $\lambda_1$, $\lambda_2$, $\lambda_3$ and $\lambda_4$ are respectively set to 0.3, 1.0, 0.01 and 1.5. The temperature coefficient $\tau_c$ for contrastive learning is set to 0.7.
The parameters of the baseline methods were set exactly as in the original publications. 

\subsection{Performance Comparation}
\noindent\textbf{Clustering performance.} 
To evaluate the performance of scAGC, we employ two widely-used evaluation metrics, the Normalised Mutual Information (NMI) \cite{strehl2002cluster} and the Adjusted Rand Index (ARI) \cite{hubert1985comparing}. The higher the value of the metrics, the better the clustering performance.
The clustering performance of our method compared to the baseline methods on 9 scRNA-seq datasets is presented in Table \ref{tab:cluster_result}. Each clustering method was run ten times to compute the average, and the values highlighted represent the best results. 

As shown in Table \ref{tab:cluster_result}, on the 9 datasets, our method yields the best 9 and 7 NMI and ARI scores compared to other baseline algorithms, respectively. Even in "QX\_LM", the NMI and ARI reached 0.9509 and 0.9719. 
Traditional clustering methods exhibit instability across diverse datasets due to their reliance on raw input features. In complex single-cell datasets, such features often fail to capture the underlying structure or salient patterns, leading to suboptimal and inconsistent clustering performance.
Meanwhile, our analysis reveals that deep embedding approaches do not consistently improve clustering performance. This limitation can be attributed to the reliance solely on gene expression data, which may inadequately capture the complex and sparse nature of scRNA-seq data. 
Compared to deep graph embedding clustering methods, our proposed scAGC consistently achieves superior performance. This suggests that approaches relying on fixed cell graphs struggle to capture meaningful structure and often hinder effective GNN training, particularly under long-tailed degree distributions commonly observed in single-cell data. Although scE2EGAE employs end-to-end cell graph construction, it lacks task-specific optimization, which leads to suboptimal performance in clustering tasks.
In summary, scAGC can perform better than other methods.
\begin{figure}[!t]
    \centering
    \includegraphics[width=1.00\linewidth]{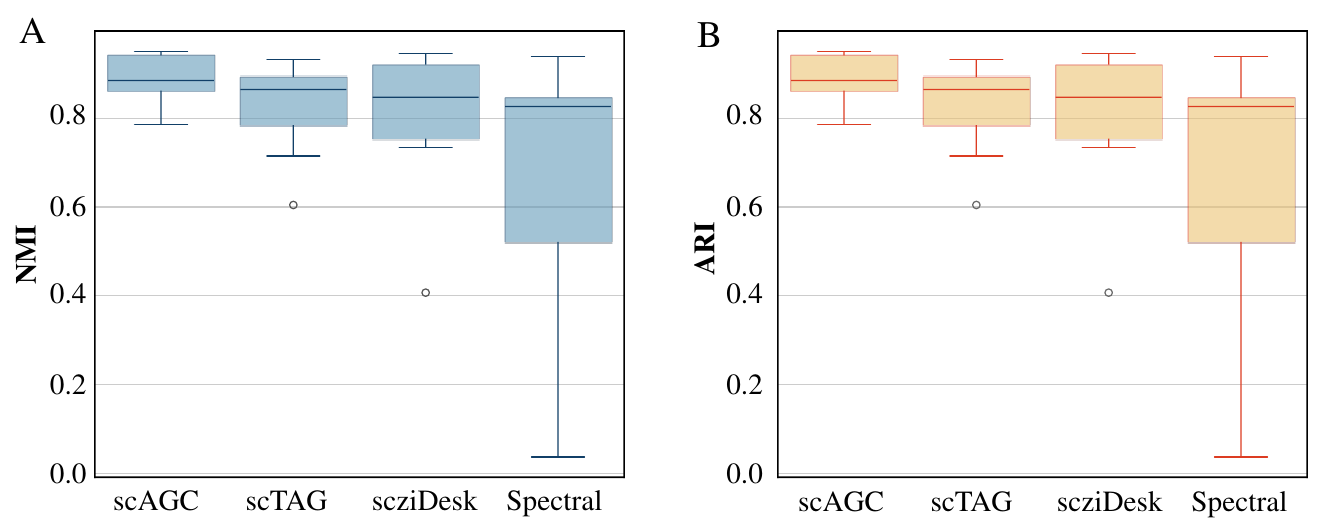}
    \caption{Comparison of the robustness of different methods across differents scRNA-seq datasets.}
    \label{fig:robust_degree}
\end{figure}
\begin{figure}[!t]
    \centering
    \includegraphics[width=1.00\linewidth]{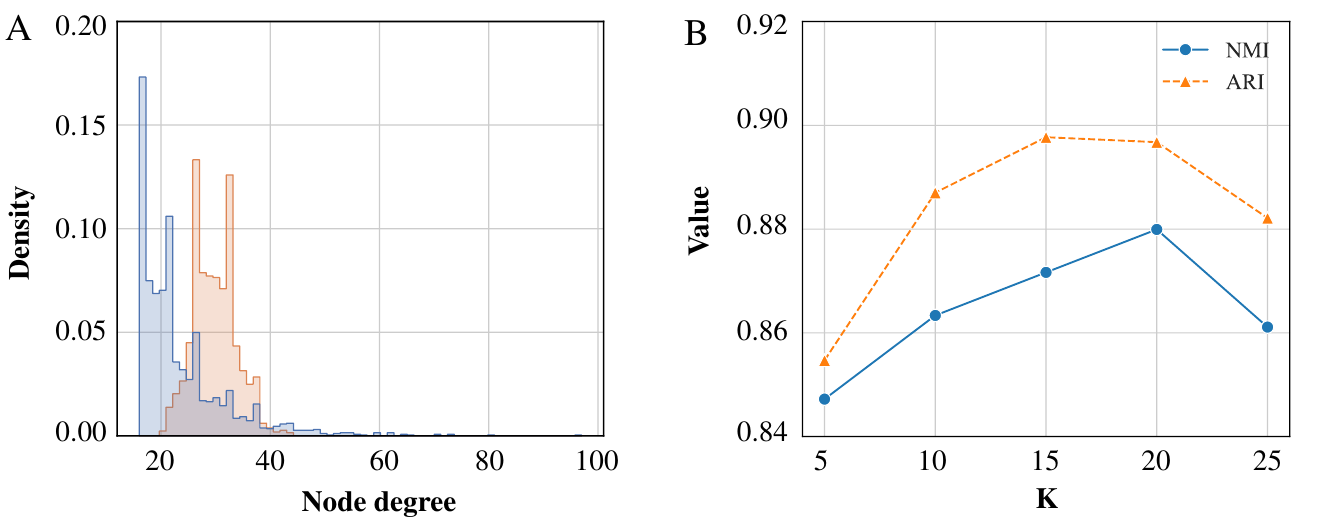}
    \caption{(A) Comparison of degree distributions between the KNN-constructed graph and the end-to-end optimized adaptive graph. (B) Comparison of the average ARI and NMI values with different numbers of genes.}
    \label{fig:long_tail}
\end{figure}

\begin{figure*}[!t]
    \centering
    \includegraphics[width=1.0\linewidth]{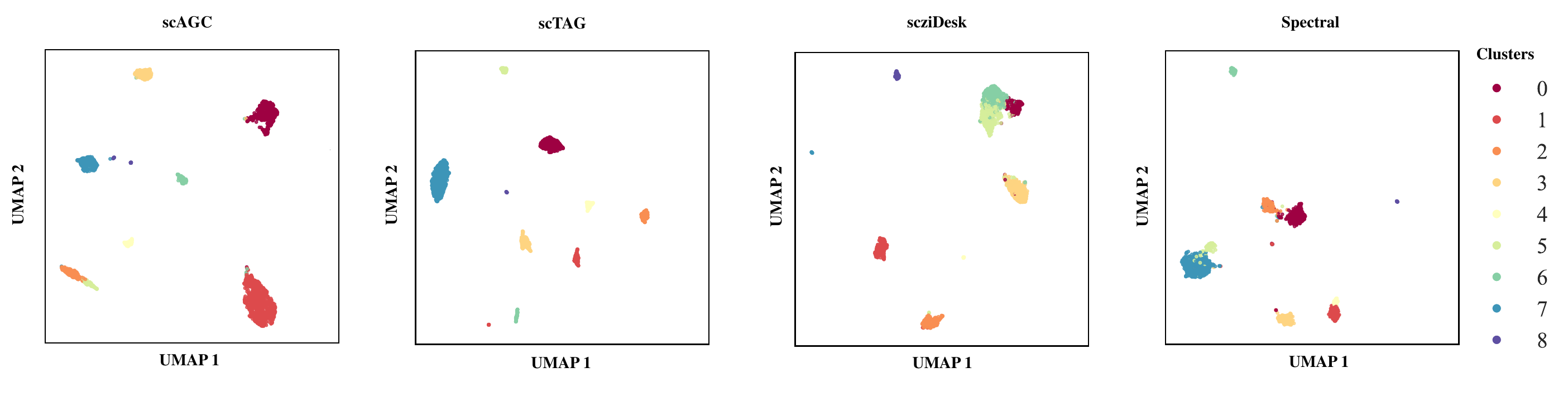}
    \caption{Comparison of clustering results with 2D visualization by UMAP on the Muraro dataset.}
    \label{fig:results_visual}
\end{figure*}
To evaluate the robustness of our method across datasets, we compare it with a representative baseline using box plots as shown in Fig. \ref{fig:robust_degree}. The results show that our method exhibits significantly lower performance variance across multiple datasets, highlighting its stability and further demonstrating its overall superiority.
Moreover, to demonstrate the intuitive discrimination ability of scAGC, we employ the Uniform Manifold Approximation and Projection (UMAP) technique with default parameters to project the scRNA-seq data into a 2D space on the Muraro datasets in Fig. \ref{fig:results_visual}. The resulting visualization clearly shows well-separated clusters, further validating the effectiveness of our method.

\noindent\textbf{Mitigation of the Topological Long-tail Distribution.} 
To compare the quality of end-to-end adaptive graphs and KNN-based graphs, we visualize their degree distributions on the Muraro dataset as shown in Fig. \ref{fig:long_tail} (A). The KNN graph follows a long-tailed distribution, where most nodes have low degrees, but a few super nodes emerge with very high degrees. These nodes aggregate excessive information and broadcast to many neighbors, leading to feature dilution and increased risk of over-smoothing, which weakens cluster boundaries.
In contrast, the adaptive graph exhibits a bell-shaped degree distribution, with most nodes having degrees between 25 and 40. This results in more balanced and localized message passing, where each node receives similar amounts of information, effectively reducing feature oversmoothing and improving clustering performance.

\subsection{Parameter Analysis}
\noindent \textbf{Impact of the neighbor parameter $K$.} 

When constructing the cell graph using the KNN algorithm, the parameter $K$ determines the number of neighbors connected to each node. To assess the impact of this parameter, we evaluated the model using $K$ values of 5, 10, 15, 20, and 25. As shown in Fig.~\ref{fig:long_tail} (B), both NMI and ARI scores increase rapidly from $K = 5$ to $K = 10$, with ARI peaking at $K = 15$ and NMI reaching its highest value at $K = 20$. However, ARI performance starts to decline once $K$ exceeds 15. Based on these observations, we set $K = 15$ as the default value in our subsequent experiments.

\begin{figure}[!t]
    \centering
    \includegraphics[width=1.0\linewidth]{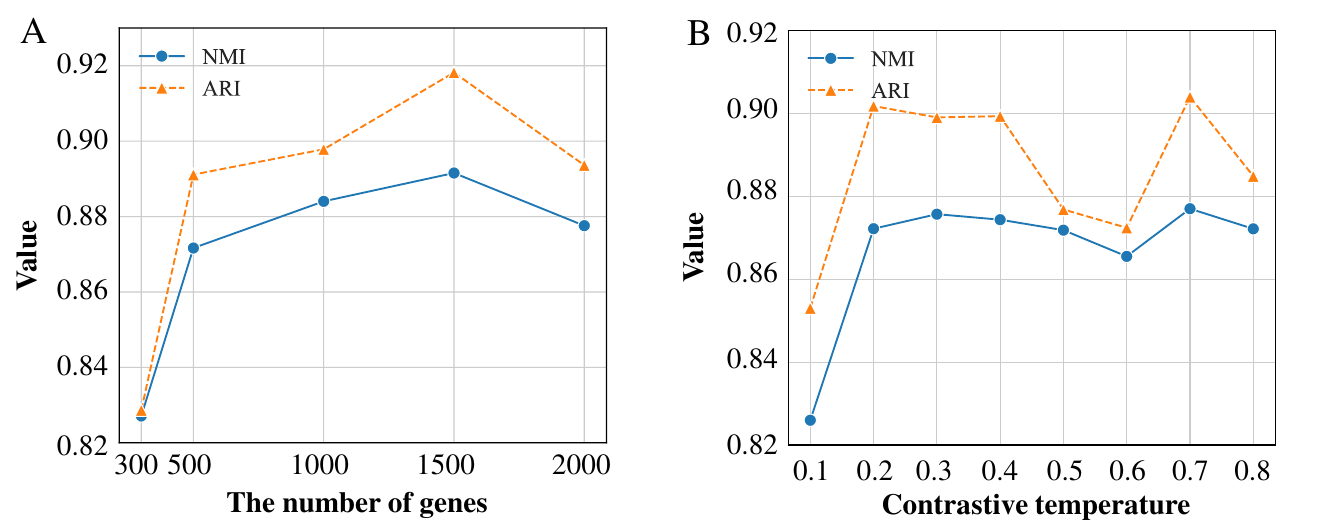}
    \caption{(A) Comparison of the average ARI and NMI values with different numbers of genes. (B) Comparison of the average ARI and NMI values with different contrastive temperature parameters.}
    \label{fig:para_genes_temper}
\end{figure}
\begin{table}[!t]
    \caption{Ablation study measured by NMI and ARI values.}
    \centering
    \begin{tabular}{c|cc}
        \toprule
            \makebox[0.05\textwidth][c]{Methods}& \makebox[0.1\textwidth][c]{NMI} & \makebox[0.05\textwidth][c]{ARI} \\
        \midrule
        scAGC w/o $\mathcal{L}_{cg}$ & 0.8772 & 0.8982 \\
        scAGC w/o adaptive graphs & 0.6346 & 0.6360 \\
        scAGC w/o adaptive graphs, $\mathcal{L}_{cg}$  & 0.7396 & 0.7434 \\
        \cmidrule{1-3}
         scAGC & 0.8951 & 0.9193 \\
       \bottomrule
    \end{tabular}
    \label{tab:ablation}
\end{table}

\noindent\textbf{Different numbers of variable genes analysis.} 

In single-cell RNA-seq analysis, highly variable genes are essential for capturing cell-type specificity and preserving biological relevance. To investigate the impact of the number of highly variable genes on model performance, we evaluated our method on nine datasets, each processed using 300, 500, 1000, and 1500 selected genes. As shown in Fig.~\ref{fig:para_genes_temper} (B), both average NMI and ARI scores generally improve with more genes, reaching optimal performance when 1500 highly variable genes are used. Based on these results, we set the number of highly variable genes to 1500 in all subsequent experiments.

\noindent\textbf{Impact of the contrastive temperature parameter $\tau_{c}$.} 
In this study, we employ the temperature parameter in contrastive learning to control the magnitude of structural perturbations, thereby preventing abrupt changes in the adjacency matrix. To evaluate the impact of this hyperparameter, we vary the temperature in the range of $\{0.1,0.2,0.3,0.4,0.5,0.6,0.7,0.8\}$ and report the model's performance in terms of NMI and ARI across different settings, as shown in Fig. \ref{fig:para_genes_temper}.
Results show that performance peaks at a temperature of 0.7. Lower temperatures (0.2–0.4) produce sparser and more stable graphs with clearer adjacency patterns, facilitating faster convergence. Higher temperatures (0.7–0.8) yield smoother graphs and enhanced semantic consistency. In contrast, intermediate values (0.5–0.6) lead to performance drops, possibly due to unstable optimization or insufficient structural cues.
Considering the overall performance trade-off, we select 0.7 as the temperature parameter for contrastive learning in our final configuration.

\subsection{Ablation Study}
In this experiment, we conduct an ablation study to evaluate the contribution of each component of our method across 9 scRNA-seq datasets. Specifically, we test the following variants: 
1) Removing the contrastive learning guidance loss;
2) Removing adaptive graph construction and instead applying contrastive training with perturbed static graphs;
3) Removing both modules simultaneously.
As shown in Table~\ref{tab:ablation}, incorporating multiple views consistently improves clustering performance. Notably, the combined use of adaptive graph construction and contrastive guidance yields the most significant performance gains, highlighting the importance of jointly modeling scRNA-seq data through adaptive cell graphs with contrastive guidance.

\section{Conclusion}
In this paper, we propose scAGC, a single-cell clustering framework designed to address the unique challenges of scRNA-seq data. To overcome the limitations of fixed graphs, scAGC employs a topology-adaptive graph autoencoder with differentiable Gumbel-Softmax sampling, enabling end-to-end optimization of both graph structure and cell representations. This dynamic construction mitigates long-tailed degree distributions and reduces feature over-smoothing.
To capture the sparse and overdispersed nature of gene expression, scAGC integrates a Zero-Inflated Negative Binomial loss for robust reconstruction. A contrastive objective further stabilizes graph evolution, while clustering is guided by a KL divergence loss. Experiments on nine real-world scRNA-seq datasets confirm the superiority of scAGC, which consistently outperforms existing state-of-the-art methods in both NMI and ARI.

\bibliographystyle{IEEEtran}
\bibliography{bibm25}

\end{document}